\pdfoutput=1
\documentclass[11pt]{article}

\usepackage[final]{acl}

\usepackage{times}
\usepackage{latexsym}

\usepackage[T1]{fontenc}
\usepackage[utf8]{inputenc}

\usepackage{microtype}

\usepackage{inconsolata}

\usepackage{graphicx}
\usepackage{booktabs}
\usepackage{amsmath}
\usepackage{amssymb}
\usepackage{multirow,multicol}
\usepackage{subfigure}

\title{HGAdapter: Hypergraph-based Adapters in Language Models for \\Code Summarization and Clone Detection}

\author{
 Guang Yang$^{1,2}$\thanks{\ Guang Yang is the corresponding author. E-mail: \texttt{51215901104@stu.ecnu.edu.cn} or \texttt{yangguang8@gtht.com}\\
 Code is available at \href{https://github.com/qiankunmu/HGAdapter}{https://github.com/qiankunmu/HGAdapter}}, 
 Yujie Zhu$^{2}$
\\
 $^{1}$Data Technology Group, Technology Research and Development Department, \\Guotai Haitong Securities, China\\
 $^{2}$School of Computer Science and Technology, East China Normal University, China
\\
\texttt{\{51215901104, 52205901006\}@stu.ecnu.edu.cn}
}

\begin{document}
\maketitle
\begin{abstract}
Pre-trained language models (PLMs) are increasingly being applied to code-related tasks. 
Although PLMs have achieved good results, they do not take into account potential high-order data correlations within the code. 
We propose three types of high-order correlations in code tokens, i.e. abstract syntax tree family correlation, lexical correlation, and line correlation. 
We design a tokens and hyperedges generator to capture these high-order data correlations. 
We improve the architecture of hypergraph neural networks and combine it with adapter tuning to propose a novel hypergraph-based adapter (HGAdapter) to fine-tune PLMs. 
HGAdapter can encode high-order data correlations and is allowed to be inserted into various PLMs to enhance performance. 
Experiments were conducted on several public datasets, including six languages of code summarization and code clone detection tasks. 
Our methods improved the performance of PLMs in datasets to varying degrees. 
Experimental results validate the introduction of high-order data correlations that contribute to improved effectiveness. 
\end{abstract}

\section{Introduction}
In recent years, the emergence of pre-trained language models (PLMs), large language models (LLMs), and small language models (SLMs) has driven the application of associated techniques to code-related tasks, including code clone detection, code classification, code summarization, and more~\cite{NiuL0022,zhang2024unifying}. 
Unlike natural language, code inherently contains richer structural information. 
Consequently, in code-related tasks, numerous language models incorporate syntactic and semantic code structures to enhance task performance~\cite{FengGTDFGS0LJZ20,GuoRLFT0ZDSFTDC21,GuoLDW0022}. 
However, the high-order data correlation inherent in code remains underexplored in current language models. 
High-order data correlation refers to the relationship among multiple entities as a unified unit, usually involving more than two entities, distinguishing it from a pairwise relationship~\cite{Feng2018HypergraphNN,Berge1973GraphsAH}. 
Current language models mainly employ Transformer-based encoder or decoder architectures, where the self-attention mechanism considers pairwise relationships between all tokens. 
%Yet, they fail to extract features that treat a group of tokens as a cohesive unit. 
However, this framework lacks the capability to capture cohesive group-level features where multiple tokens can be treated as an integrated feature unit. 
Our research reveals that in the source code, there exist high-order data correlations that enable the extraction of such integrated feature units. 

\begin{figure}[htbp]
    \centering
    \includegraphics[width=\columnwidth]{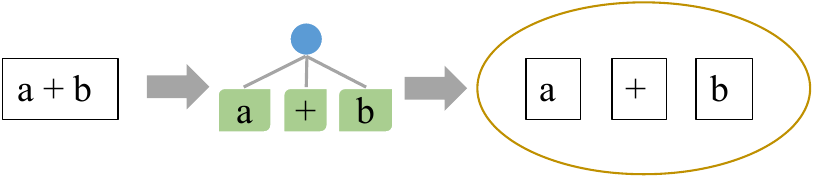}
    \caption{AST family correlation. The tokens `a', `+', and `b' belong to the same AST parent node, forming an addition operation, and can thus be treated as a whole unit.}
    \label{figast}
\end{figure}

The first category of these correlations originates from the cohesive relationships among syntax tree nodes that share a common parent node in the abstract syntax tree (AST).
The code can be parsed into an AST which represents the structure of the code. 
The text of the code corresponds to the leaf nodes of the AST.
Tokens that belong to the same AST parent node can have their features computed as a unified whole to capture structural semantics, indicating that they together form a specific code structure. 
We call it AST family correlation. 
An example of AST family correlation is shown in Figure \ref{figast}. 

\begin{figure}[htbp]
    \centering
    \includegraphics[width=\columnwidth]{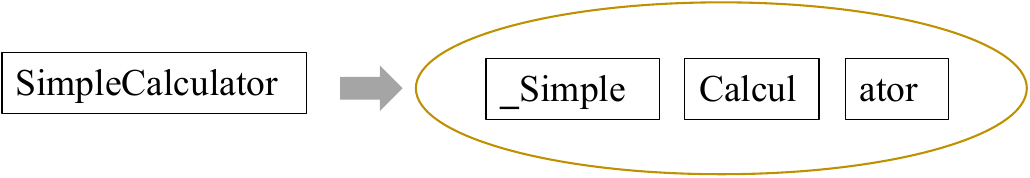}
    \caption{Lexical correlation. We use Llama BBPE-based tokenizer for tokenization. SimpleCalculator is a class name that was split into three tokens. These tokens can be viewed as a whole to extract features from their original class name.}
    \label{figlex}
\end{figure}

The second category are correlations within lexical units. 
In code, there are numerous long lexical constructs, such as function names, variable names, and class names, which frequently comprise multiple semantic components concatenated via camel case or underscore delimiters. 
When undergoing tokenization, these unified lexical units may be fragmented into discrete tokens. 
The resulting split tokens can have their features computed as a whole to preserve the characteristics of their parent lexical unit. 
We call this lexical correlation.
An example of lexical correlation is shown in Figure \ref{figlex}. 

\begin{figure}[htbp]
    \centering
    \includegraphics[scale=0.45]{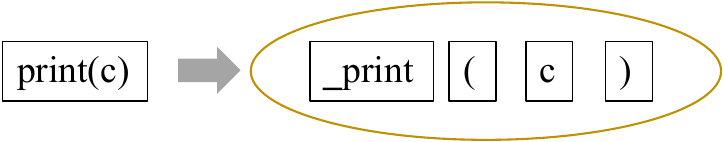}
    \caption{Line correlation. These tokens can be treated as a line to extract features. }
    \label{figline}
\end{figure}

The third category involves tokens within the same line of code. 
Programmers typically write code with lines as the basic unit. 
Tokens on the same line of original code can be processed as a whole, providing additional granularity to represent code organization patterns. 
We call it line correlation. 
An example of line correlation is shown in Figure \ref{figline}. 
To extract the three types of high-order data correlation from the code, we design a tokens and hyperedges generator. 

Existing language models only consider tokens as pairwise relationships, not as collective units for the above high-order correlation feature encoding. 
To address this gap, we propose a novel hypergraph-based adapter (HGAdapter). 
Hypergraphs are the standard framework for representing high-order correlations. 
Unlike the edges in a standard graph, which only connect two entities, the edges in a hypergraph can connect any number of entities called the hyperedge~\cite{Berge1973GraphsAH}. 
Hypergraph neural networks (HGNNs) have emerged as the most widely used neural networks to capture high-order data correlations~\cite{Feng2018HypergraphNN,Yadati2018HyperGCNAN,Kim2020HypergraphAN,Huang2021UniGNNAU}. 
Adapters are lightweight modules inserted into pre-trained models for fine-tuning~\cite{Rebuffi2017LearningMV,houlsby19a}. 
We improved the two-stage architecture of HGNNs and combined it with adapters to propose HGAdapter, we implement it as an adapter module inserted into language models, enabling the extraction and computation of the high-order features to enhance language models performance. 

We conduct experiments on code summarization, a generation task, and code clone detection, an understanding task.
We used public datasets of the two tasks to evaluate our method. 
Experimental results validate that our method improves the performance of language models, with the introduction of high-order data correlations contributing to an improvement in effectiveness. 

The main contributions of this paper are: 
\begin{itemize} 
\item We propose to introduce high-order data correlations within code into PLMs. 
We propose AST family correlation, lexical correlation, and line correlation. 
We propose a tokens and hyperedges generator to capture these high-order data correlations. 
\item We propose a novel HGAdapter to encode high-order data correlations in PLMs that is improved from the architecture of hypergraph neural networks and combined with adapter tuning. 
\item We conducted experiments on public datasets of code summarization and code clone detection to validate our method. 
\end{itemize}

\section{Related Work}\label{related}
\subsection{Code Summarization and Clone Detection}
Code summarization is a task that takes the code as input and outputs a description of that code, which can be considered as a generation task. 
Code clone detection is a task that takes two code samples as input, determines whether the two code fragments are functionally equivalent, and can be considered as a binary classification task. 
In the early stages, most approaches designed different neural networks to extract code structural features for such tasks~\cite{alon2018codeseq,Chen2019CodeSW,WangLM0J20}. 
Following the success of PLMs such as BERT~\cite{Devlin2019BERTPO} and GPT~\cite{Radford2018ImprovingLU} in natural language processing, many researchers have developed PLMs for code-related tasks such as CodeBERT~\cite{FengGTDFGS0LJZ20},  CodeT5~\cite{Wang2021CodeT5IU}. 
Compared to natural language, code has more structural features, so some PLMs will consider structural information of code, such as GraphCodeBERT~\cite{GuoRLFT0ZDSFTDC21} incorporates data flows and UniXcoder~\cite{GuoLDW0022} learn the text of AST. 
Currently, due to the success of LLMs, some LLMs for code have emerged, including StarCoder~\cite{starcoder}, Code Llama~\cite{Rozire2023CodeLO}, DeepSeek series~\cite{deepseekv3}, Qwen Coder series~\cite{qwen2.5coder,qwen3}, and more. 

These PLMs are typically built on the Transformer~\cite{Vaswani2017AAN} encoder-only, decoder-only, or encoder-decoder architecture~\cite{NiuL0022,zhang2024unifying}. 
For code summarization, encoder-only PLMs require training a decoder to generate results in a seq2seq manner, whereas decoder-only models can directly produce output. 
In code clone detection with encoder-only models, the hidden state vector corresponding to the starting position of the PLM output is used as the code representation. 
Two code vectors are concatenated and fed into a fully-connected neural network classifier for binary classification. 
Our method follows this process as established in previous work. 

\subsection{Hypergraph Neural Networks}
HGNNs are proposed to encode high-order data correlations. 
Similarly to graph neural networks (GNNs), an HGNN consists of multiple layers, each layer updating the hidden vectors of the nodes. 
HGNNs employ a two-stage message passing process at each layer: aggregating messages from nodes to their connected hyperedges and aggregating messages from hyperedges to their connected nodes to update nodes vectors. 
Representative works are HGNN~\cite{Feng2018HypergraphNN}, HyperGCN~\cite{Yadati2018HyperGCNAN}, HGAN~\cite{Kim2020HypergraphAN}, 
UniGNN~\cite{Huang2021UniGNNAU}, etc.
In code-related tasks, HEAT~\cite{GeorgievBA22} and HDHGN~\cite{YangJD23} have mined different high-order data correlations in code and improved HGNNs. 

\subsection{Adapter Tuning}
Adapter tuning is a kind of parameter-efficient fine-tuning (PEFT) method that inserts small-scale parameters between PLM layers.  
During fine-tuning, the PLM parameters are frozen and not updated, while only the inserted parameters are trained. 
This approach can achieve or even surpass the performance of full fine-tuning of the PLM. 
Such methods require the storage of only a small number of parameters and reduce the resources needed for training. 
Representative works are adapter~\cite{Rebuffi2017LearningMV}, adapter for NLP~\cite{houlsby19a}, AdapterFusion~\cite{Pfeiffer2020AdapterFusionNT}, MADX~\cite{Pfeiffer2020MADXAA}, etc.
Structural adapter~\cite{RibeiroZG21,MontellaNHBR23} combines the GNNs and adapter tuning. 

\section{Methods}\label{method}
Our methodology has two core components: the tokens and hyperedges generator module and the HGAdapter module. 
Before code text is input into the language model, it needs to be tokenized. 
To capture high-order data correlation in the code, we design a tokens and hyperedges generator instead of directly tokenizing the code. 
To encode high-order data correlation, we design the HGAdapter, which is integrated as an adapter module between the PLM layers. 

\subsection{Tokens and Hyperedges Generator}\label{thig}
A tokens and hyperedges generator is designed to extract high-order data correlations within code text. 
Since we use hyperedge to represent high-order data correlation, in HGAdapter we refer to the high-order data correlation as hyperedge, like the AST family hyperedge, lexical hyperedge, and line hyperedge. 

This module includes a parser and a tokenizer. 
We use tree-sitter\footnote{https://tree-sitter.github.io/tree-sitter/} as the parser. 
We adopt the corresponding original tokenizers of different PLMs. 
When code is input, the parser will parse the code into an AST. 
We perform a postorder traversal to visit the AST. 
When visiting a leaf node, the code text of the node is first extracted and fed into the tokenizer to generate tokens. 
We assign each token a unique token id. 
If the number of tokens exceeds two, a new hyperedge id is created to map these tokens ids to the hyperedge, while recording the hyperedge type as lexical hyperedge. 
We use the COO format\footnote{https://docs.pytorch.org/docs/stable/sparse.html\#sparse-coo-docs} to record the correspondence between token ids and hyperedge ids, where each entry consists of (token\_id, hyperedge\_id) pairs representing associations. 
Similarly, when visiting the leaf node, we can also obtain the corresponding line number for the code text. 
If we encounter a new line number, we create a new hyperedge id. 
We use a dictionary to record line numbers and their corresponding hyperedge ids. 
We also record the correspondence between these split token ids and the hyperedge id as the line hyperedge. 
Upon completing visiting the leaf node, we return token sequences, token ids, hyperedge ids, and hyperedge types to its parent node. 

When visiting a parent node, obtain all tokens returned by its leaf nodes. 
If the number of tokens exceeds two, create a new hyperedge id, record these token ids as belonging to this hyperedge, and label it as an AST family hyperedge. 
Return all tokens, token ids, hyperedge ids, and hyperedge types, including those collected from its child nodes to the higher-level parent node. 

After completing the AST traversal, we obtain the token sequence of the code, token ids, and hyperedge ids that record hyperedge information, along with all hyperedge types. 

\subsection{Hypergraph-based Adapters} 
\begin{figure}[htbp]
    \centering
    \includegraphics[width=\columnwidth]{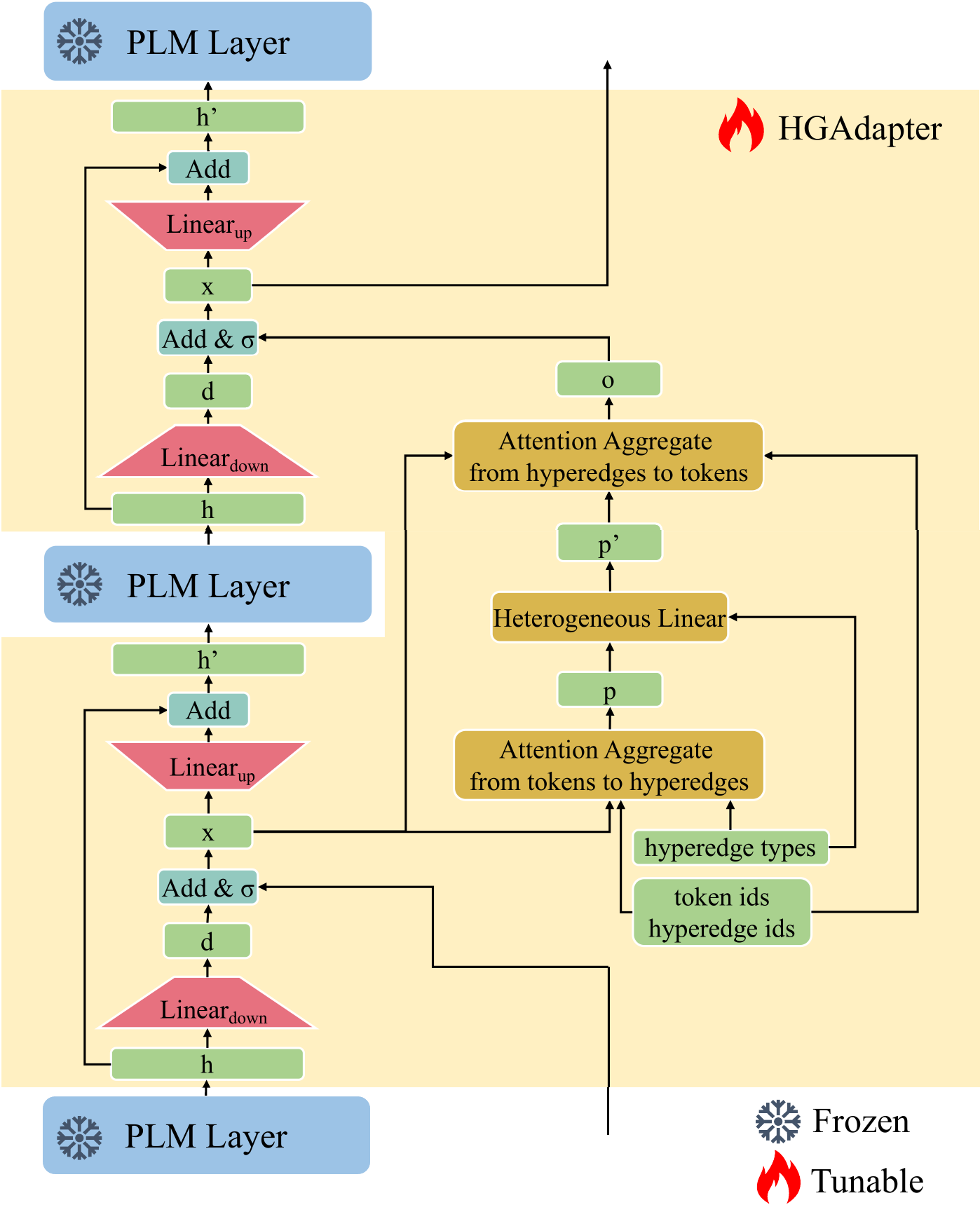}
    \caption{Overall structure of HGAdapter}
    \label{fighga}
\end{figure}

The HGAdapter is inserted between the PLM layers. 
The structure is illustrated in Figure \ref{fighga}. 
We improved the framework of HGNNs, incorporating a simplified attention mechanism, introducing heterogeneous linear transformations, and integrating it with adapters. 
During fine-tuning, all PLM parameters are frozen, and only the adapter parameters are updated. 

After input of the token sequence into PLM, the PLM layer outputs the hidden state vectors corresponding to each token, denoted as $h^l_{0},~h^l_{1},\ldots,h^l_{N-1}$, where $N$ is the total number of tokens, $l$ denotes the layer number in PLM, $l \in \{1,2,\ldots,L\}$ with $L$ representing the total number of layers, $h_{n}^l \in \mathbb{R}^{C}$, $C$ is the size of the dimension of the hidden state vector , $n = 0,~1,\ldots,N-1$ represents the token id in the sequence. 

These vectors, along with token ids, hyperedge ids, and hyperedge types, are input into the HGAdapter. 
Token ids and hyperedge ids record which tokens belong to which hyperedges. 
Here, a hyperedge is defined as $e$, and the set of tokens it has is denoted as $T(e)$. 
Given token id $n$, the set of hyperedges to which it belongs is denoted as $S(n)$. 
The type of hyperedge $e$ is denoted as $\rho(e)$. 

First, project the hidden states vectors $h_n^l$ into a lower dimension $C_{down}$ to get $d_{n}^{l}$, as shown in Equation \ref{eq1}, where $d_{n}^{l} \in \mathbb{R}^{C_{down}}$. 
\begin{equation}
d_{n}^{l}=W_{down}^{l}h_{n}^l+b_{down}^l
\label{eq1}
\end{equation}

Afterward, $d_n^l$ and the output $o_n^{l-1}$ of the HGAdapter of the previous layer are summed and passed through an activation function, which yields x, where $\sigma$ is the activation function. 
We use the ReLU activation function~\cite{GlorotBB11}. 
\begin{equation}
x_{n}^{l}=\sigma{\left(d_{n}^{l}+o_{n}^{l-1}\right)}
\label{eq2}
\end{equation}
For the first layer, $d_n^1$ is fed directly into the activation function. 
\begin{equation}
x_{n}^{1}=\sigma{\left(d_{n}^{1}\right)}
\label{eq3}
\end{equation}

Then, project $x_{n}^{l}$ from $C_{down}$ into the normal dimension $C$ and add ${h_n^l}$, as shown in Equation \ref{eq48}. 
The vectors $h{'}^l_{0}, h{'}^{l}_{1}, \ldots, h{'}^{l}_{N-1}$ are then propagated to the next layer of the PLM. 
\begin{equation}
h{'}^l_n=W^l_{up}x_{n}^{l}+b^l_{up}+h_n^l
\label{eq48}
\end{equation}

For $o_n^l$, its computation process is as follows. 
First, we aggregate messages from tokens to hyperedges. 
HGAdapter aggregates the $x^{l}$ of tokens that belong to the same hyperedge to obtain a vector $p^{l}$ representing the hyperedge. 
We employ a relatively simplified attention mechanism to aggregate $x^{l}$. 
When a token $n$ belongs to a hyperedge $e$, their attention score $\alpha_{ne}^{l}$ is calculated as in Equation \ref{eq42}, with the computation $softmax$ occurring across the tokens belonging to the same hyperedge $e$. 
The $q_{\rho(e)}$ is a query vector representing the type of hyperedge $e$, which will be updated during training. 
\begin{equation}
\alpha^l_{ne} = Softmax\left( \frac{{\left(q_{\rho(e)}\right)}^{T}x_{n}^{l}}{\sqrt{C_{down}}} \right)
\label{eq42}
\end{equation}
Aggregate $x^{l}$ to get $p^{l}_{e}$ as shown in Equation \ref{eq43}. 
\begin{equation}
p^{l}_{e} = {\sum_{n \in T(e)}{\alpha^l_{ne} x_{n}^{l}}}
\label{eq43}
\end{equation}

Later, $p^{l}_e$ is subject to a heterogeneous linear transformation that varies depending on the hyperedge type, to integrate the information of the hyperedge type into the vector, 
as shown in Equation \ref{eq44}.
\begin{equation}
p{'}_{e}^{l}=W^l_{\rho(e)}p_{e}^{l}+b^l_{\rho(e)}
\label{eq44}
\end{equation}

Lastly, aggregate the hyperedge vectors $p{'}^{l}$ to the tokens. 
The aggregation uses the same attention mechanism. 
When a hyperedge $e$ has a token $n$, their attention score $\alpha^l_{en}$ is calculated as in Equation \ref{eq45}, with the computation $softmax$ occurring across the hyperedges having the same token $n$. 
\begin{equation}
\alpha^l_{en} = Softmax\left( \frac{{\left(x_{n}^{l}\right)}^{T}p{'}_{e}^{l}}{\sqrt{C_{down}}} \right)
\label{eq45}
\end{equation}
Aggregate $p{'}^{l}$ to get the new token vector $o^{l}_{n}$ as shown in Equation \ref{eq46}. 
\begin{equation}
o^{l}_{n} = {\sum_{e \in S(n)}{\alpha^l_{en}p{'}_{e}^{l}}}
\label{eq46}
\end{equation}
The $o_n^l$ is then added to $d_n^{l+1}$ of the next layer for subsequent operations, as in Equation \ref{eq2}. 

\section{Experiment Settings}\label{es}
% We provide our implementation of the source code in \href{https://github.com/qiankunmu/HGAdapter}{https://github.com/qiankunmu/HGAdapter}. 

\subsection{Datasets}
\begin{table*}[bhtp]
    \centering
    \small
    \begin{tabular}{c c c c c c c}
    % \toprule
    \hline
         & Ruby&JavaScript&Java&Go&PHP&Python\\
    % \midrule
    \hline     Training&24927&58025&164923&167288&241241&251820 \\ 
         Validation&1400&3885&5183&7325&12982&13914\\
         Testing&1261&3291&10955&8122&14014&14918\\
         Avg. tokens&126.27 &173.35 &155.82 &133.23 &271.02 & 280.58\\
         Avg. hyperedges&53.27 &74.45 &67.29 &55.86 &71.24 & 73.65\\
    % \bottomrule
    \hline
    \end{tabular}
    \caption{Statistics of CodeSearchNet.}
    \label{tab2}
\end{table*}
\begin{table}[htbp]
    \centering
    \small
    \begin{tabular}{c c}
    % \toprule
    \hline
         & BigCloneBench\\
    % \midrule
    \hline
         Training&901028 \\ 
         Validation&415416 \\
         Testing&415416 \\
         % Positive samples in train&450862\\
         % Negative samples in train& 450166\\
         % Positive samples in valid& 53839\\
         % Negative samples in valid& 361577\\
         % Positive samples in test& 56820\\
         % Negative samples in test& 358596\\
         Avg. tokens& 401.84\\
         Avg. hyperedges& 182.11\\
         Language&Java\\
    % \bottomrule
    \hline
    \end{tabular}
    \caption{Statistics of BigCloneBench.}
    \label{tab1}
\end{table}
For code summarization, we use a widely applied public dataset CodeSearchNet\cite{codesearchnet} which contains a million functions collected from open-source code. 
It has datasets in 6 languages: Ruby, JavaScript, Java, Go, PHP, and Python. 
The version of the dataset we used is filtered and organized provided by CodeXGLUE\cite{LuGRHSBCDJTLZSZ21}. 
The statistics is shown in Table \ref{tab2}, where Avg.tokens means the average number of tokens per code snippet, and Avg.hyperedges means the average number of hyperedges per code snippet. 
Each sample has a code snippet and a segment of natural language description. 

For code clone detection, we use the public dataset BigCloneBench\cite{svajlenko2014towards} which is the most widely used Java code clone detection dataset. 
The version of the dataset we used is also provided by CodeXGLUE\cite{LuGRHSBCDJTLZSZ21}. 
%which was cleaned and formatted consistent with Wang et al.\cite{WangLM0J20}. 
The statistical information of the dataset is presented in Table \ref{tab1}. 
Each sample has two code snippets for clone detection. 

\subsection{Baselines}
In code summarization, we select several language models that have demonstrated outstanding performance in this generation task as baselines, including RoBERTa~\cite{Liu2019RoBERTaAR}, CodeBERT~\cite{FengGTDFGS0LJZ20}, GraphCodeBERT~\cite{GuoRLFT0ZDSFTDC21}, UniXcoder~\cite{GuoLDW0022}, Code Llama 7B~\cite{Rozire2023CodeLO}, TinyLlama-Math\&Code~\cite{tinyllama}, Qwen2.5-Coder-0.5B~\cite{qwen2.5coder} as comparisons. 

In code clone detection, we select several language models that have demonstrated outstanding performance in this understanding task as baselines, including CodeBERT~\cite{FengGTDFGS0LJZ20}, GraphCodeBERT~\cite{GuoRLFT0ZDSFTDC21}, and UniXcoder~\cite{GuoLDW0022}. 

We used full fine-tuning or adapter tuning~\cite{houlsby19a} for comparative experiments. 
We also used the structural adapter~\cite{RibeiroZG21,MontellaNHBR23} for comparison. 
For the structural adapter, we treat AST family, lexical, and line correlations as pairwise token relationships rather than hyperedges, and we employ the GNN the same as the structural adapter. 

\subsection{Training Settings}
We implement the HGAdapter based on the PyTorch\footnote{https://pytorch.org/docs/stable/index.html}, transformers\footnote{https://huggingface.co/docs/transformers/index}, and adapters\footnote{https://docs.adapterhub.ml/index.html} libraries. 
We conducted the experiments on a machine that has 64GB of RAM and an RTX 3090 GPU with 24GB. 
We select the tree-sitter\footnote{https://tree-sitter.github.io/tree-sitter/} to parse the code snippets into ASTs. 
We use the tokenizers used by the corresponding PLMs. 
The PLMs and tokenizers that we used are all provided by the official on Hugging Face\footnote{https://huggingface.co/}. 
The hyperparameter configuration of the PLMs remains unchanged. 
The dimension size of the vectors in the adapter is 64. 
We employ the cross-entropy loss function in training. 

In code summarization, we select BLEU-4~\cite{bleu} as the evaluation metric. 
We use the BLEU-4 evaluation implemented by Hugging Face\footnote{https://huggingface.co/spaces/evaluate-metric/bleu}.
We use Adam optimizer\cite{Kingma2015Adam} with a learning rate of $1 \times 10^{-4}$. 
The batch size is 64. 
The number of training epochs is 20. 
The adapter parameters with the highest BLEU-4 score in the validation set are saved. 

For code clone detection, we select F1, precision, and recall as the evaluation metric. 
We use AdamW optimizer\cite{LoshchilovHAdamW} with a learning rate of $5 \times 10^{-5}$. 
The batch size is 4 (including 8 code snippets) and the number of training epochs is 10. 
The adapter parameters with the highest F1 score in the validation set are saved. 
\begin{table*}[htbp]
    \centering
    \small
    \begin{tabular}{lccccccc}
    % \toprule
    \hline
         Models & Ruby & JavaScript & Java & Go & PHP & Python & Overall\\
    % \midrule
    \hline
        RoBERTa (Full Fine-tuning) & 11.73 & 11.88 & 16.52 & 16.49 & 21.68 & 17.14 & 15.91\\
        RoBERTa (Adapter) & 11.66 & 11.95 & 16.58 & 16.41 & 21.89 & 17.10 & 15.93\\
        RoBERTa (Structural Adapter) & 12.45 & 12.71 & 17.01 & 16.99 & 22.32 & 17.85 & 16.56 \\
        RoBERTa (HGAdapter) & \textbf{14.27} & \textbf{14.60} & \textbf{19.06} & \textbf{18.55} & \textbf{24.03} & \textbf{18.94} & \textbf{18.24}\\
    \hline
        CodeBERT (Full Fine-tuning) & 12.13 & 13.85 & 17.68 & 16.58 & 22.87 & 18.09 & 16.87\\
        CodeBERT (Adapter) & 12.02 & 13.74 & 17.51 & 16.59 & 22.90 & 17.92 & 16.78\\
        CodeBERT (Structural Adapter) & 12.93 & 14.26 & 18.14 & 17.06 & 23.27 & 18.37 & 17.34 \\
        CodeBERT (HGAdapter) & \textbf{14.31} & \textbf{16.14} & \textbf{19.70} & \textbf{18.86} & \textbf{24.62} & \textbf{19.52} & \textbf{18.86}\\
    % \midrule
    \hline
        GraphCodeBERT (Full Fine-tuning) & 12.42 & 14.80 & 18.98 & 17.86 & 24.02 & 18.05 & 17.69\\
        GraphCodeBERT (Adapter) & 12.51 & 14.75 & 19.00 & 17.63 & 23.94 & 18.06 & 17.65\\
        GraphCodeBERT (Structural Adapter) & 13.06 & 15.33 & 19.47 & 18.34 & 24.63 & 18.45 & 18.21 \\
        GraphCodeBERT (HGAdapter) & \textbf{14.95} & \textbf{16.94} & \textbf{21.06} & \textbf{19.82} & \textbf{25.68} & \textbf{19.77} & \textbf{19.70}\\
    \hline
        UniXcoder (Full Fine-tuning) & 14.93 & 15.73 & 20.01 & 19.07 & 25.96 & 19.18 & 19.15\\
        UniXcoder (Adapter) & 15.04 & 15.76 & 19.78 & 18.73 & 25.87 & 19.09 & 19.05\\
        UniXcoder (Structural Adapter) & 15.51 & 16.22 & 20.24 & 19.17 & 26.28 & 19.46 & 19.48 \\
        UniXcoder (HGAdapter) & \textbf{16.92} & \textbf{18.07} & \textbf{22.04} & \textbf{20.95} & \textbf{27.81} & \textbf{20.57} & \textbf{21.06}\\
    \hline
        Code Llama 7B (Adapter) & 15.46 & 17.03 & 21.26 & 19.56 & 27.06 & 20.09 & 20.08\\
        Code Llama 7B (Structural Adapter) & 16.03 & 17.45 & 21.70 & 19.97 & 27.45 & 20.51 & 20.52 \\
        Code Llama 7B (HGAdapter) & \textbf{17.14} & \textbf{18.62} & \textbf{22.86} & \textbf{21.22} & \textbf{28.00} &\textbf{21.34}& \textbf{21.53}\\
    % \bottomrule
    \hline
        TinyLlama-Math\&Code(Full Fine-tuning) & 14.96 & 16.19 & 19.08 & 18.21 & 23.87 & 18.15 & 18.41 \\
        TinyLlama-Math\&Code(Adapter) & 14.78 & 16.03 & 19.35 & 18.11 & 23.81 & 18.09 & 18.36 \\
        TinyLlama-Math\&Code(Structural Adapter) & 15.12 & 16.45 & 19.63 & 18.46 & 24.07 & 18.36 & 18.68 \\
        TinyLlama-Math\&Code(HGAdapter) & \textbf{16.89} & \textbf{18.04} & \textbf{21.11} & \textbf{20.19} & \textbf{25.54} & \textbf{19.92} & \textbf{20.28} \\
    \hline
        Qwen2.5-Coder-0.5B (Full Fine-tuning) & 15.10 & 16.32 & 19.61 & 18.76 & 25.97 & 19.43 & 19.20 \\
        Qwen2.5-Coder-0.5B (Adapter) & 14.97 & 16.12 & 19.54 & 18.83 & 26.01 & 19.22 & 19.12 \\
        Qwen2.5-Coder-0.5B (Structural Adapter) & 15.29 & 16.41 & 19.75 & 19.03 & 26.18 & 19.34 & 19.33 \\
        Qwen2.5-Coder-0.5B (HGAdapter) & \textbf{17.03} & \textbf{18.17} & \textbf{21.31} & \textbf{20.56} & \textbf{27.73} & \textbf{20.97} & \textbf{20.96} \\
    \hline
    \end{tabular}
    \caption{BLEU-4 results of code summarization on CodeSearchNet (\%)}
    \label{tab4}
\end{table*}
\section{Results}\label{r}
\subsection{Code Summarization}
The BLEU-4 scores of the models in the testing sets are shown in Table \ref{tab4}. 
As we can see, the HGAdapter has improved the performance of PLMs to varying degrees in most programming languages. 
Compared to full fine-tuned PLMs, HGAdapter demonstrates significant overall improvements, specifically showing performance gains of 2.33 over RoBERTa, 1.99 over CodeBERT, 2.01 over GraphCodeBERT, 1.91 over UniXcoder, 1.87 over TinyLlama-Math\&Code and 1.76 over Qwen2.5-Coder-0.5B. 
HGAdapter also achieves notable improvements over general adapter-tuned PLMs, demonstrating performance gains of 2.31 with RoBERTa, 2.08 with CodeBERT, 2.05 with GraphCodeBERT, 2.01 with UniXcoder, 1.45 with Code Llama 7B, 1.92 with TinyLlama-Math\&Code, and 1.84 with Qwen2.5-Coder-0.5B. 
These results validate that the introduction of high-order data correlations can improve the effectiveness of code summarization, and our HGAdapter can encode high-order data correlations within language models and enhance their performance. 

HGAdapter shows advantages over the structural adapter that also incorporate structural information, surpassing its performance by 1.68 on RoBERTa, 1.52 on CodeBERT, 1.49 on GraphCodeBERT, 1.58 on UniXcoder, 1.01 on Code Llama 7B, 1.60 on TinyLlama-Math\&Code, and 1.63 on Qwen2.5-Coder-0.5B. 
This validates that for AST family, lexical and line structural information, extracting features by treating tokens as high-order data correlations is better than only as pairwise relationships. 

Comparative analysis reveals that HGAdapter shows more substantial improvements over adapter in RoBERTa, CodeBERT, and GraphCodeBERT, while showing relatively smaller performance gains in Code Llama 7B, TinyLlama-Math\&Code, and Qwen2.5-Coder-0.5B. 
This is likely because the former are language models based on the Transformer encoder architecture, while the latter are decoder-based language models. 
Encoder-based language models excel at extracting richer contextual information, and HGAdapter further extracts features from their hidden vectors based on high-order data correlations, thus achieving greater improvement. 
We can observe that HGAdapter provides limited improvement for Code Llama 7B, probably because its large parameter size already grants it strong code feature extraction capabilities, leaving less room for HGAdapter to enhance its performance. 

We also observed that HGAdapter achieves relatively greater improvements on the Ruby and JavaScript datasets, likely because these datasets are relatively smaller in scale. 
As a result, full fine-tuning or general adapter tuning may yield insufficient training effectiveness. 
HGAdapter compensates for this limitation by providing richer feature information. 

\subsection{Code Clone Detection}
The results of the models on the testing set are shown in Table \ref{tab3}. 
The results show that HGAdapter improves precision, recall, and F1 scores for most language models. 
Compared to full fine-tuned PLMs, HGAdapter improves F1 scores by 1.28 for CodeBERT, 1.23 for GraphCodeBERT, and 1.12 for UniXcoder. 
Compared to adapter tuned PLMs, HGAdapter achieves F1 score improvements of 1.31 for CodeBERT, 1.21 for GraphCodeBERT and 1.17 for UniXcoder. 
These results validate that introducing high-order data correlations through the HGAdapter can indeed enhance the understanding of code by PLMs, thereby improving the performance on the code clone detection. 

HGAdapter outperforms the structural adapter with an F1 score of 1.01 in CodeBERT, 0.80 in GraphCodeBERT, and 0.89 in UniXcoder. 
This also indicates that for code clone detection task, incorporating AST family, lexical, and line structural information as high-order correlations is better than treating them only as pairwise relationships. 

For CodeBERT, HGAdapter achieves improvements of 1.87 in precision and 0.69 in recall compared to full fine-tuning. 
In GraphCodeBERT, HGAdapter shows performance improvements of 1.76 in precision and 0.71 in recall. 
For both PLMs, the HGAdapter makes greater improvements in precision. 
For UniXcoder, HGAdapter achieves a significant 2.35 recall improvement, but does not make precision higher. 
This observation may come from the UniXcoder inherently high-precision baseline, where HGAdapter recall enhancement comes at the cost of slight precision degradation. 
\begin{table}[htbp]
    \centering
    \scriptsize
    \begin{tabular}{lccc}
    % \toprule
    \hline
         Models & Precision & Recall & F1\\
    % \midrule
    \hline
         CodeBERT (Full Fine-tuning) & 94.76 & 94.72 & 94.74\\
         CodeBERT (Adapter) & 94.78 & 94.64 & 94.71\\
         CodeBERT (Structural Adapter) & 95.05 & 94.98 & 95.01\\
         CodeBERT (HGAdapter) & \textbf{96.63} & \textbf{95.41} & \textbf{96.02}\\
    \hline
         GraphCodeBERT (Full Fine-tuning) & 95.39 & 94.68 & 95.03\\
         GraphCodeBERT (Adapter) & 95.44 & 94.65 & 95.05\\
         GraphCodeBERT (Structural Adapter) & 96.01 & 94.92 & 95.46\\
         GraphCodeBERT (HGAdapter) & \textbf{97.15} & \textbf{95.39} & \textbf{96.26}\\
    \hline
         UniXcoder (Full Fine-tuning) & \textbf{97.26} & 92.84 & 95.01\\
         UniXcoder (Adapter) & 97.15 & 92.87 & 94.96\\
         UniXcoder (Structural Adapter) & 97.11 & 93.45 & 95.24\\
         UniXcoder (HGAdapter) & 97.08 & \textbf{95.19} & \textbf{96.13}\\
    \hline
    \end{tabular}
    \caption{Results of code clone detection on BigCloneBench (\%)}
    \label{tab3}
\end{table}

\subsection{Ablation Study}
\begin{figure}[t]
    \centering
    \subfigure[Results of CodeBERT on Ruby, JavaScript and Java]{
        \includegraphics[width=\columnwidth]{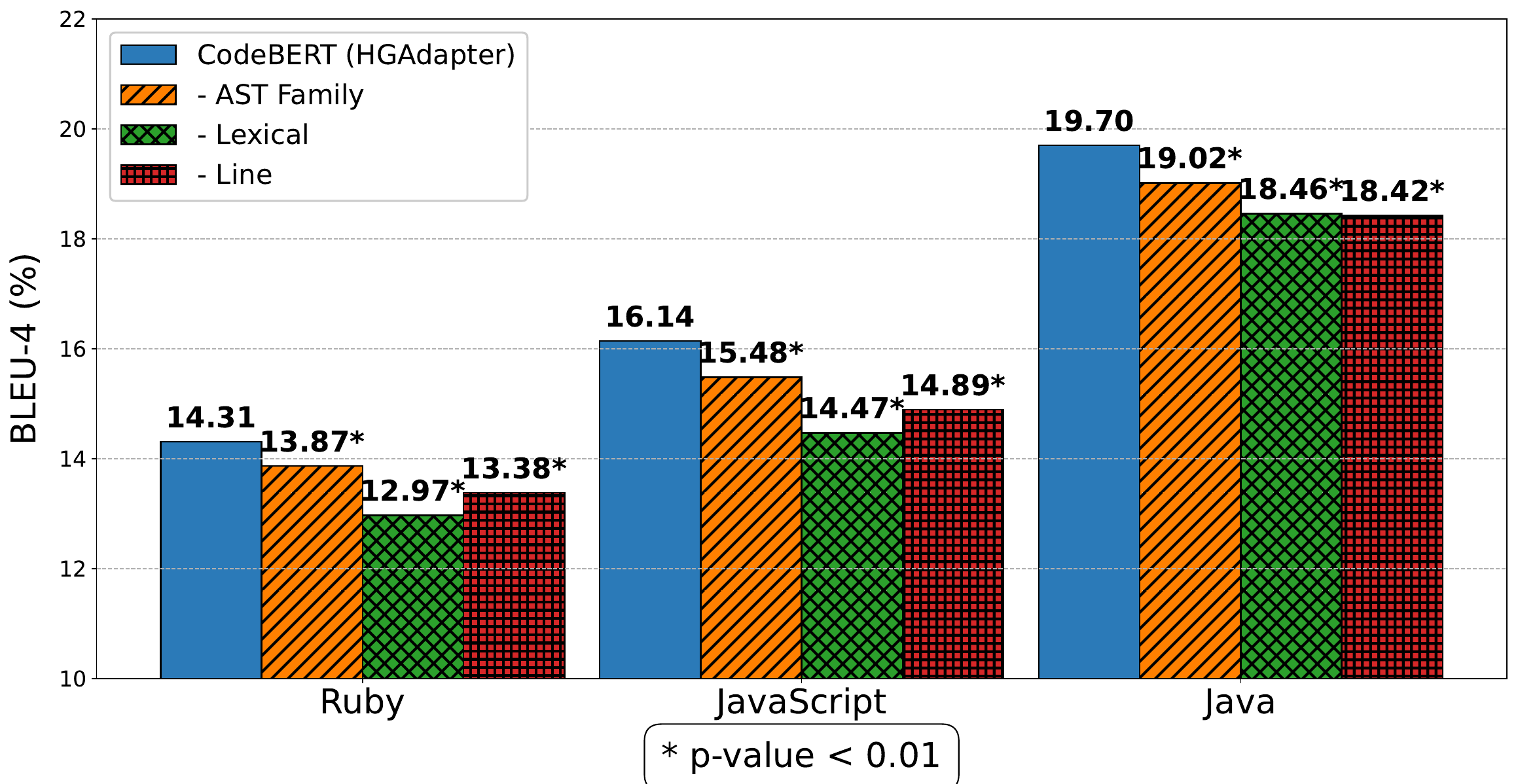}
        \label{figcsa:sub1}
    }
    \subfigure[Results of CodeBERT on Go, PHP and Python]{
        \includegraphics[width=\columnwidth]{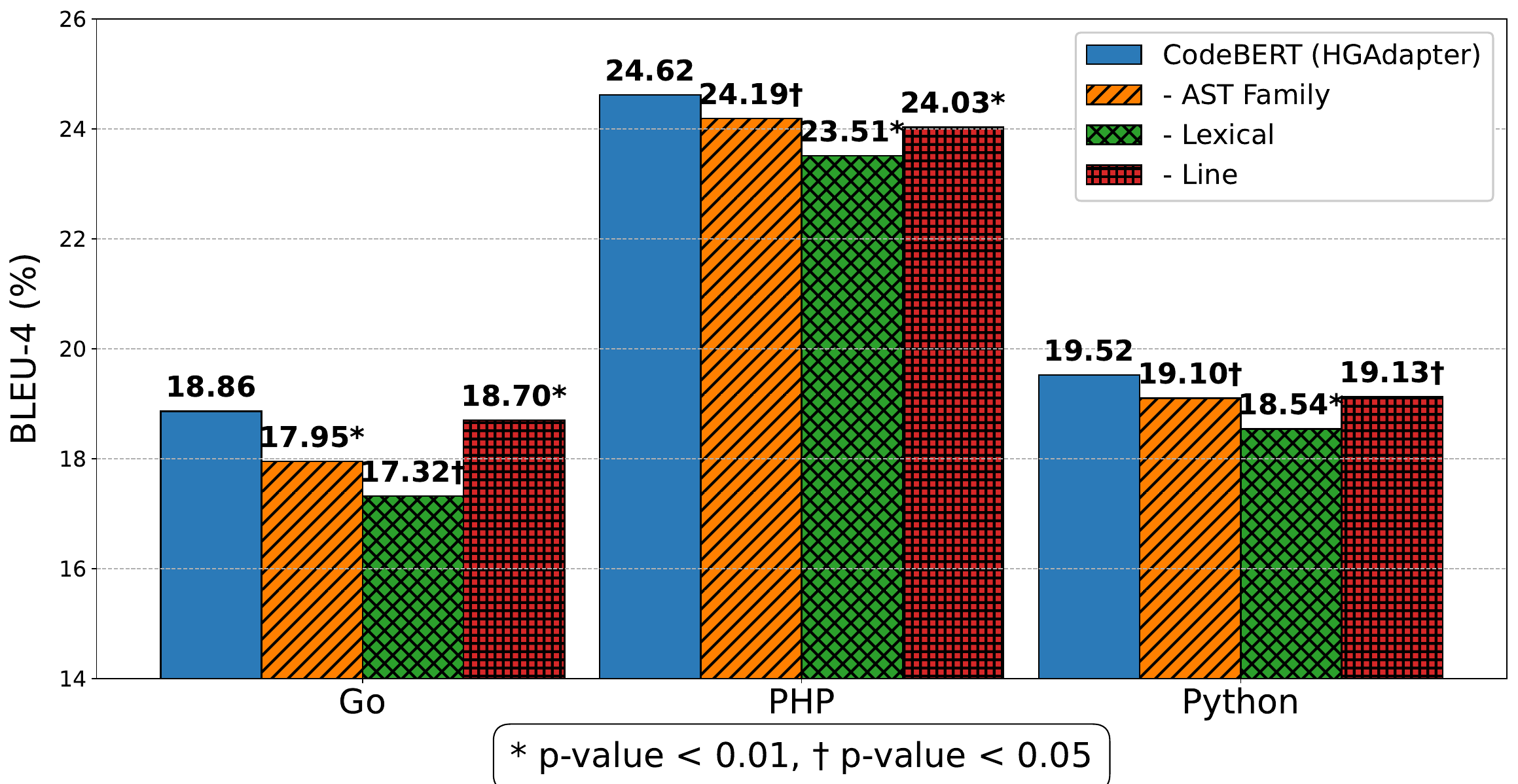}
        \label{figcsa:sub2}
    }
    \caption{Results of ablation study on code summarization in CodeBERT (\%)}
    \label{figcsa}
\end{figure}

\begin{figure}[t]
    \centering
    \subfigure[Results of TinyLlama-Math\&Code on Ruby, JavaScript and Java]{
        \includegraphics[width=\columnwidth]{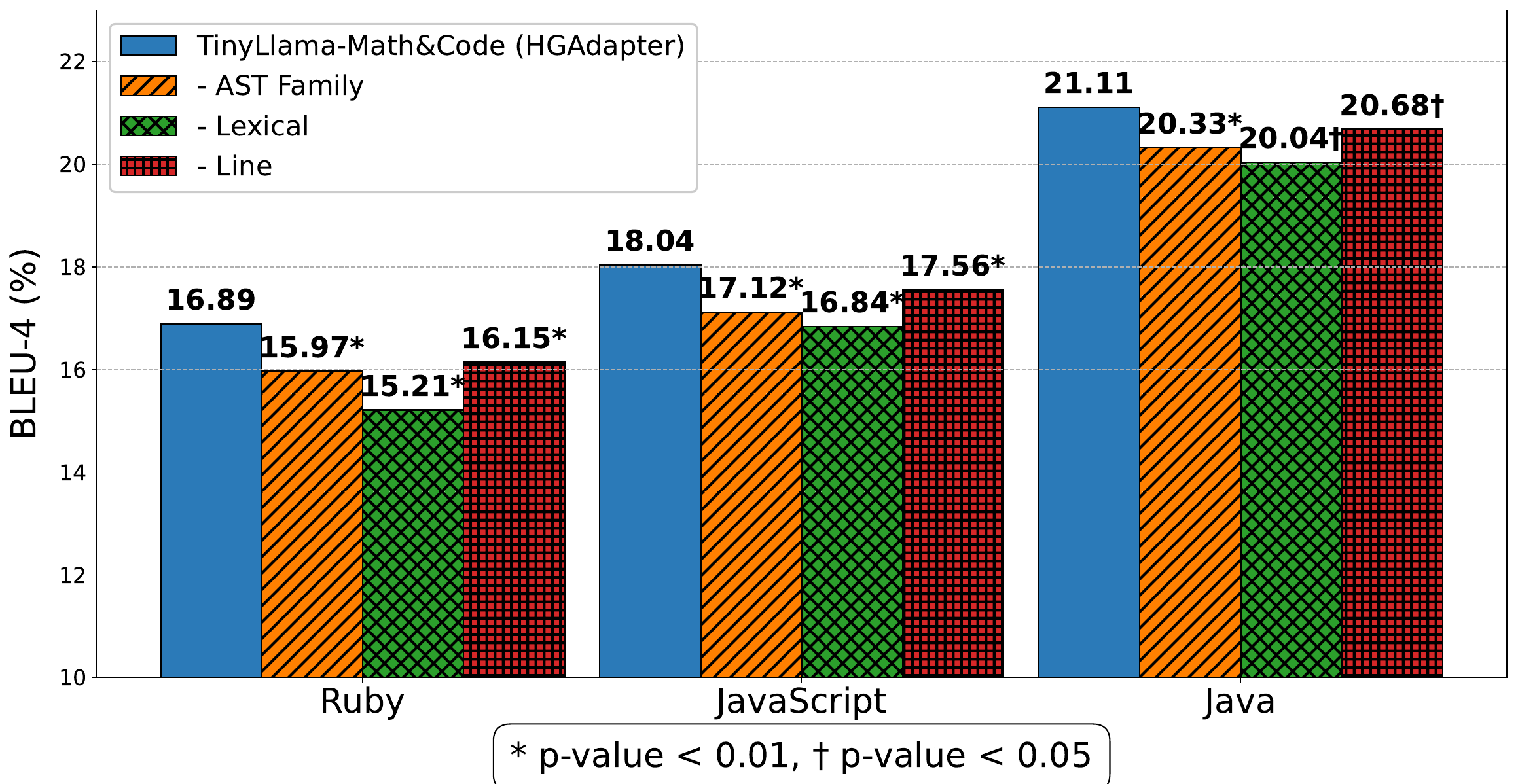}
        \label{figcsb:sub1}
    }
    \subfigure[Results of TinyLlama-Math\&Code on Go, PHP and Python]{
        \includegraphics[width=\columnwidth]{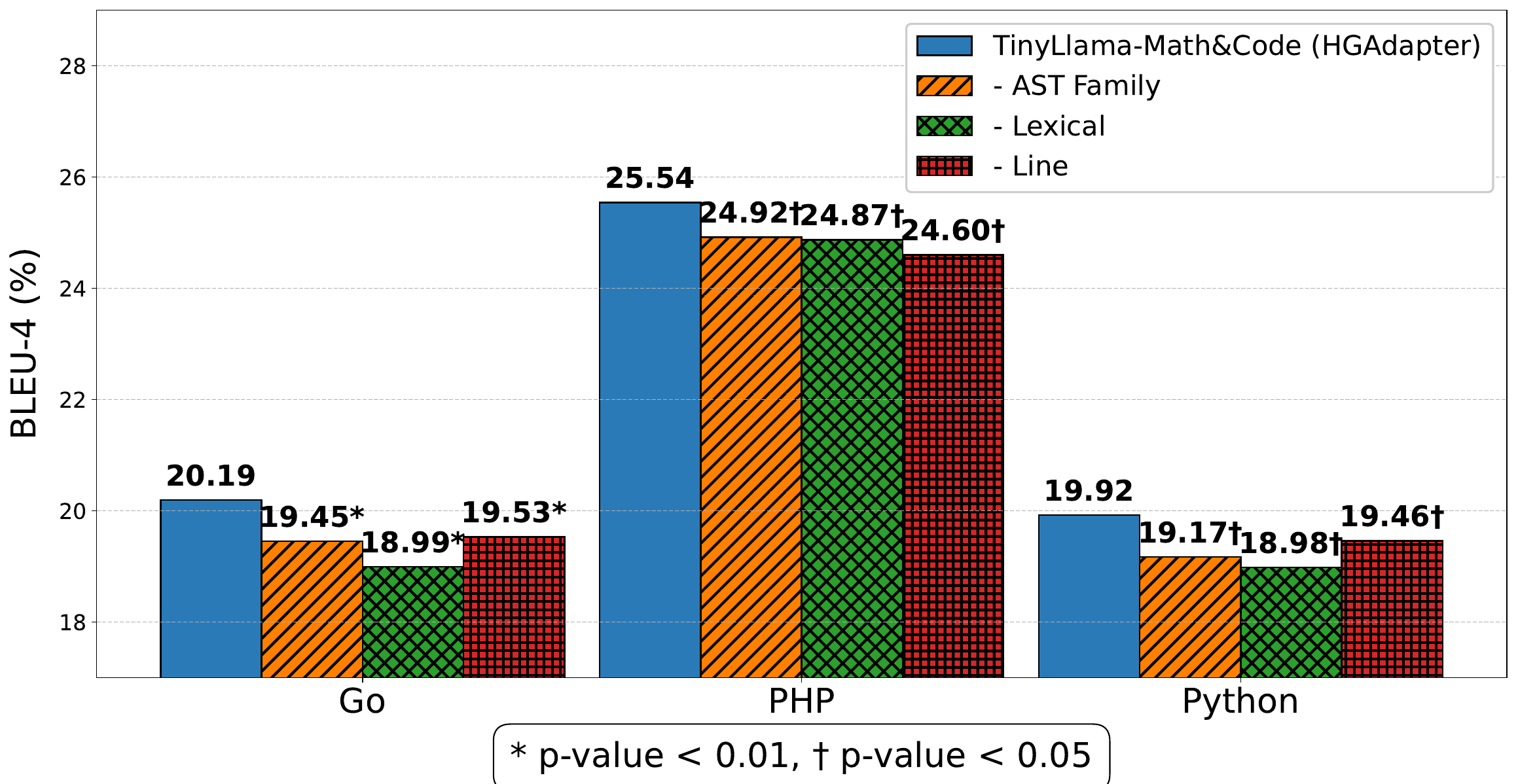}
        \label{figcsb:sub2}
    }
    \caption{Results of ablation study on code summarization in TinyLlama-Math\&Code (\%)}
    \label{figcsb}
\end{figure}

In code summarization, we perform ablation experiments in CodeBERT to validate that the introduction of the three types of high-order data correlations can improve performance.  
They correspond to the AST family hyperedge, lexical hyperedge, and line hyperedge. 
We separately ablate each type of hyperedge in HGAdapter. 
The results are shown in Figure \ref{figcsa} and Figure \ref{figcsb}.
In CodeBERT, removing AST family hyperedges, lexical hyperedges, and line hyperedges results in an average decrease of 0.59, 1.31, and 0.78 respectively. 
On TinyLlama-Math\&Code, removing AST family hyperedges, lexical hyperedges, and line hyperedges resulted in overall performance drops of 0.79, 1.12, and 0.62 respectively.
Through multiple experiments and comparisons with HGAdapter, the resulting p-value was less than 0.05. 
The experimental results validate that these three types of high-order data correlations can improve the performance of language models in code summarization. 
We also found that lexical hyperedges have a more significant impact compared to the other two types. 
%18.27 17.55 18.08
%19.49 19.16 19.66
\begin{figure}[tb]
    \centering
    \includegraphics[width=\columnwidth]{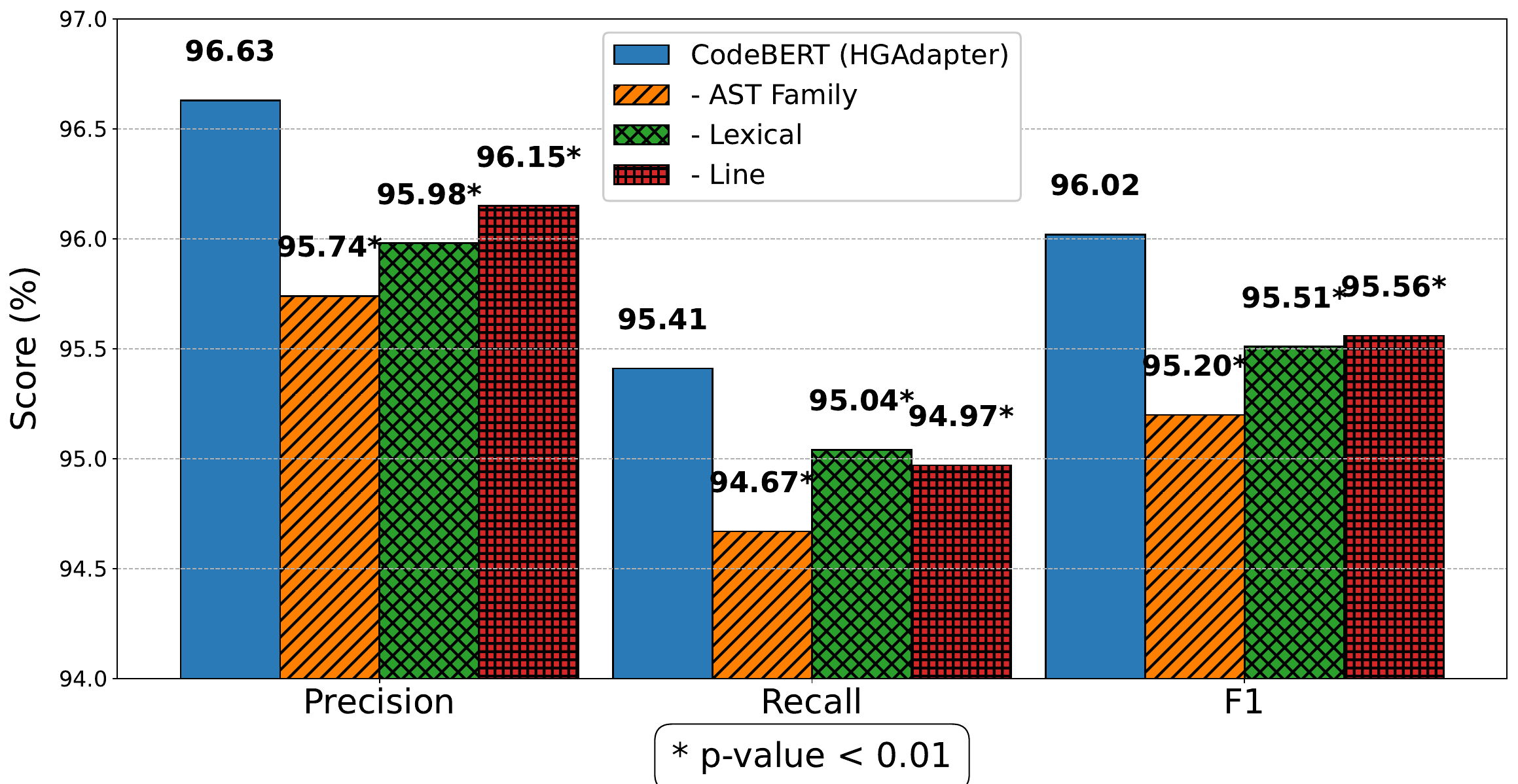}
    \caption{Results of ablation study on code clone detection (\%)}
    \label{figclonea}
\end{figure}
\begin{table*}[htbp]
    \centering
    \small
    \begin{tabular}{lccc}
        \hline
        PLM name & Params & Adapter Params & HGAdapter Params \\
        \hline
        RoBERTa, CodeBERT, GraphCodeBERT & 125M & 1.2M & 1.3M \\
        UniXcoder& 126M & 1.2M & 1.3M\\
        Code Llama 7B & 6.7B & 16.9M & 17.3M \\
        TinyLlama-Math\&Code & 1.1B & 5.8M & 6.1M \\
        Qwen2.5-Coder-0.5B & 0.5B & 2.8M & 3.1M \\
        \hline
    \end{tabular}
    \caption{Number of parameters for different PLMs and adapters}
    \label{tab5}
\end{table*}

In code clone detection, we also conduct ablation experiments on CodeBERT in BigCloneBench to validate that the introduction of the three types of high-order data correlations can improve performance. 
We separately ablate each type of hyperedge in HGAdapter. 
The results are shown in Figure \ref{figclonea}.
% \begin{table}[htbp]
%     \centering
%     \small
%     \begin{tabular}{lcccc}
%     % \toprule
%     \hline
%          & Precision & Recall & F1 & p-value\\
%     % \midrule
%     \hline
%         CodeBERT (HGAdapter) & 96.63 & 95.41 & 96.02 &\\
%         $-$~AST family hyperedge & 95.74 & 94.67 & 95.20&<0.001\\
%         $-$~lexical hyperedge & 95.98 & 95.04 & 95.51&<0.001\\
%         $-$~line hyperedge & 96.15 & 94.97 & 95.56&<0.001\\
%     % \bottomrule
%     \hline
%     \end{tabular}
%     \caption{Results of ablation study on BigCloneBench (\%)}
%     \label{tab5}
% \end{table}
It can be seen that removal of AST family hyperedges, lexical hyperedges, and line hyperedges decreased performance to varying degrees. 
Removal of AST family hyperedges resulted in a 0.89 decrease in precision, a 0.74 decrease in recall, and a 0.82 decrease in F1. 
Eliminating lexical hyperedges led to a 0.65 decrease in precision, a 0.37 decrease in recall, and a 0.51 decrease in F1. 
Removing line hyperedges reduced precision by 0.48, recall by 0.44, and F1 by 0.46. 
Multiple experiments demonstrated statistically significant results that the p-value was less than 0.01 compared to the baseline without the removal of the hyperedge. 
These results validate that by introducing three types of high-order data correlations, the ability of PLMs to understand code can be enhanced from different perspectives. 
The results also indicate that AST family hyperedges have a more substantial impact on performance compared to other types. 
This is likely because the AST family high-order correlations enable language models to better comprehend the code structure, leading to superior performance in program understanding tasks. 

\subsection{Number of Parameters}
The number of parameters for different PLMs and their corresponding inserted adapters, as well as the HGAdapter, are shown in Table \ref{tab5}, where 1M represents 1 million and 1B represents 1 billion. 
The dimension size of the hidden vectors in the adapter is 64. 
Among them, RoBERTa, CodeBERT and GraphCodeBERT all have the same number of parameters. 

Compared to RoBERTa, CodeBERT, GraphCodeBERT and UniXcoder, the parameter count of HGAdapter is only 1\% of theirs. 
Compared to Code Llama 7B, the parameter count of HGAdapter is even as low as 0.3\% of its. 
Compared to TinyLlama-Math\&Code, the parameter size of HGAdapter is only 0.5\% of its. 
Similarly, compared to Qwen2.5-Coder-0.5B, the parameter count of the HGAdapter is merely 0.6\% of its. 
We can observe that, compared to PLM, the HGAdapter parameters account for only about 0.3\%-1\%. 
% This indicates that HGAdapter itself has a very small number of parameters. 

In RoBERTa, CodeBERT, GraphCodeBERT, and UniXcoder, HGAdapter increases the parameter count by approximately 8\% compared to the adapter. 
In Code Llama 7B, the HGAdapter increases the number of parameters by only 2\% compared to the adapter. 
In TinyLlama-Math\&Code, the increase is 5\%. 
In Qwen2.5-Coder-0.5B, HGAdapter increases the parameter count by 11\%. 
We find that, compared to the standard adapter, the HGAdapter introduces only around 3\%-11\% additional parameters. 
HGAdapter achieves performance improvements and does not significantly increase the number of parameters. 
This, to some extent, validates the efficiency of HGAdapter. 

\section{Conclusion}
In this paper, we propose to introduce high-order data correlations within code tokens into language models. 
We propose AST family correlation, lexical correlation, and line correlation. 
We design a tokens and hyperedges generator to capture the three types of high-order data correlation in the code. 
We improve the architecture of HGNNs and combine it with adapters to propose HGAdapter, it can encode high-order data correlations, and it is allowed to be inserted into various PLMs. 
We perform experiments on public datasets of code summarization and code clone detection tasks. 
Experimental results show that our method increases the performance of PLMs, with the introduction of high-order data correlations contributing to an improvement in results. 
Further ablation studies and parameter comparisons further validate the effectiveness of our method. 
In the future, we will explore the introduction of more high-order data correlations, the integration of more effective parameter fine-tuning methods with hypergraphs, and the application of our approach to a wider range of code-related tasks. 

\section*{Limitations}
% Although we have proposed three types of code high-order data correlations and propose HGAdapter to encode them, and have achieved promising results on code summarization and code clone detection tasks, there are still several limitations to consider. 
The impact of HGAdapter on larger-scale PLMs remains to be explored in future work. 
% LLMs with massive parameter scales and vast training datasets leave limited improvement potential for HGAdapter to achieve performance gains.
% Compared to LLMs, the HGAdapter is better suited for SLMs. 
HGAdapter is a lightweight module that introduces minimal overhead in terms of both parameters and GPU memory. 
However, due to the additional computational steps involved in hypergraph construction and processing, HGAdapter increased the training time and inference latency. 
HGAdapter relies on tasks that require complete code input and is therefore not directly applicable to other tasks without code input, such as code generation. 
% Therefore, how to achieve this with incomplete code is an important future research direction for us. 
More correlations can be mined in code or in natural language. 
Additionally, other PEFT methods could be explored by investigating how to integrate them either with HGAdapter or with high-order data correlations. 

\section*{Acknowledgements}
This research was carried out independently. 
This research did not receive any specific grant from funding agencies in the public, commercial, or not-for-profit sectors. 
Thanks to anonymous reviewers for their insightful suggestions and comments. 

\bibliography{ref}

\begin{thebibliography}{39}
\providecommand{\natexlab}[1]{#1}

\bibitem[{Alon et~al.(2019)Alon, Brody, Levy, and Yahav}]{alon2018codeseq}
Uri Alon, Shaked Brody, Omer Levy, and Eran Yahav. 2019.
\newblock code2seq: Generating sequences from structured representations of code.
\newblock In \emph{The 7th International Conference on Learning Representations, {ICLR} 2019}.

\bibitem[{Berge(1973)}]{Berge1973GraphsAH}
Claude Berge. 1973.
\newblock \emph{Graphs and hypergraphs}.

\bibitem[{Chen et~al.(2019)Chen, Hu, and Liu}]{Chen2019CodeSW}
Qiuyuan Chen, Han Hu, and Zhaoyi Liu. 2019.
\newblock Code summarization with abstract syntax tree.
\newblock In \emph{Neural Information Processing - 26th International Conference, {ICONIP} 2019}, volume 1143 of \emph{Communications in Computer and Information Science}, pages 652--660.

\bibitem[{DeepSeek{-}AI et~al.(2024)DeepSeek{-}AI, Liu, Feng, Xue, Wang, Wu, Lu, Zhao, Deng, Zhang, Ruan, Dai, Guo, Yang, Chen, Ji, Li, Lin, Dai, Luo, Hao, Chen, Li, Zhang, Bao, Xu, Wang, Zhang, Ding, Xin, Gao, Li, Qu, Cai, Liang, Guo, Ni, Li, Wang, Chen, Chen, Yuan, Qiu, Li, Song, Dong, Hu, Gao, Guan, Huang, Yu, Wang, Zhang, Xu, Xia, Zhao, Wang, Zhang, Li, Wang, Zhang, Zhang, Tang, Li, Tian, Huang, Wang, Zhang, Wang, Zhu, Chen, Du, Chen, Jin, Ge, Zhang, Pan, Wang, Xu, Zhang, Chen, Li, Lu, Zhou, Chen, Wu, Ye, Ma, Wang, Zhou, Yu, Zhou, Pan, Wang, Yun, Pei, Sun, Xiao, and Zeng}]{deepseekv3}
DeepSeek{-}AI, Aixin Liu, Bei Feng, Bing Xue, Bingxuan Wang, Bochao Wu, Chengda Lu, Chenggang Zhao, Chengqi Deng, Chenyu Zhang, Chong Ruan, Damai Dai, Daya Guo, Dejian Yang, Deli Chen, Dongjie Ji, Erhang Li, Fangyun Lin, Fucong Dai, and 80 others. 2024.
\newblock Deepseek-v3 technical report.
\newblock \emph{CoRR}, abs/2412.19437.

\bibitem[{Devlin et~al.(2019)Devlin, Chang, Lee, and Toutanova}]{Devlin2019BERTPO}
Jacob Devlin, Ming-Wei Chang, Kenton Lee, and Kristina Toutanova. 2019.
\newblock Bert: Pre-training of deep bidirectional transformers for language understanding.
\newblock In \emph{Proceedings of the 2019 Conference of the North American Chapter of the Association for Computational Linguistics: Human Language Technologies, {NAACL-HLT} 2019}, pages 4171--4186.

\bibitem[{Feng et~al.(2019)Feng, You, Zhang, Ji, and Gao}]{Feng2018HypergraphNN}
Yifan Feng, Haoxuan You, Zizhao Zhang, R.~Ji, and Yue Gao. 2019.
\newblock Hypergraph neural networks.
\newblock In \emph{The 33rd {AAAI} Conference on Artificial Intelligence, {AAAI} 2019}, pages 3558--3565.

\bibitem[{Feng et~al.(2020)Feng, Guo, Tang, Duan, Feng, Gong, Shou, Qin, Liu, Jiang, and Zhou}]{FengGTDFGS0LJZ20}
Zhangyin Feng, Daya Guo, Duyu Tang, Nan Duan, Xiaocheng Feng, Ming Gong, Linjun Shou, Bing Qin, Ting Liu, Daxin Jiang, and Ming Zhou. 2020.
\newblock Codebert: {A} pre-trained model for programming and natural languages.
\newblock In \emph{Findings of the Association for Computational Linguistics: {EMNLP} 2020}, pages 1536--1547.

\bibitem[{Georgiev et~al.(2022)Georgiev, Brockschmidt, and Allamanis}]{GeorgievBA22}
Dobrik Georgiev, Marc Brockschmidt, and Miltiadis Allamanis. 2022.
\newblock {HEAT:} hyperedge attention networks.
\newblock \emph{Trans. Mach. Learn. Res.}, 2022.

\bibitem[{Glorot et~al.(2011)Glorot, Bordes, and Bengio}]{GlorotBB11}
Xavier Glorot, Antoine Bordes, and Yoshua Bengio. 2011.
\newblock Deep sparse rectifier neural networks.
\newblock In \emph{Proceedings of the 14th International Conference on Artificial Intelligence and Statistics, {AISTATS} 2011}, volume~15, pages 315--323.

\bibitem[{Guo et~al.(2022)Guo, Lu, Duan, Wang, Zhou, and Yin}]{GuoLDW0022}
Daya Guo, Shuai Lu, Nan Duan, Yanlin Wang, Ming Zhou, and Jian Yin. 2022.
\newblock Unixcoder: Unified cross-modal pre-training for code representation.
\newblock In \emph{Proceedings of the 60th Annual Meeting of the Association for Computational Linguistics, {ACL} 2022}, pages 7212--7225.

\bibitem[{Guo et~al.(2021)Guo, Ren, Lu, Feng, Tang, Liu, Zhou, Duan, Svyatkovskiy, Fu, Tufano, Deng, Clement, Drain, Sundaresan, Yin, Jiang, and Zhou}]{GuoRLFT0ZDSFTDC21}
Daya Guo, Shuo Ren, Shuai Lu, Zhangyin Feng, Duyu Tang, Shujie Liu, Long Zhou, Nan Duan, Alexey Svyatkovskiy, Shengyu Fu, Michele Tufano, Shao~Kun Deng, Colin~B. Clement, Dawn Drain, Neel Sundaresan, Jian Yin, Daxin Jiang, and Ming Zhou. 2021.
\newblock Graphcodebert: Pre-training code representations with data flow.
\newblock In \emph{The 9th International Conference on Learning Representations, {ICLR} 2021}.

\bibitem[{Houlsby et~al.(2019)Houlsby, Giurgiu, Jastrzebski, Morrone, De~Laroussilhe, Gesmundo, Attariyan, and Gelly}]{houlsby19a}
Neil Houlsby, Andrei Giurgiu, Stanislaw Jastrzebski, Bruna Morrone, Quentin De~Laroussilhe, Andrea Gesmundo, Mona Attariyan, and Sylvain Gelly. 2019.
\newblock Parameter-efficient transfer learning for {NLP}.
\newblock In \emph{Proceedings of the 36th International Conference on Machine Learning, {ICML} 2019}, volume~97 of \emph{Proceedings of Machine Learning Research}, pages 2790--2799.

\bibitem[{Huang and Yang(2021)}]{Huang2021UniGNNAU}
Jing Huang and Jie Yang. 2021.
\newblock Unignn: a unified framework for graph and hypergraph neural networks.
\newblock In \emph{Proceedings of the 30th International Joint Conference on Artificial Intelligence, {IJCAI} 2021}, pages 2563--2569.

\bibitem[{Hui et~al.(2024)Hui, Yang, Cui, Yang, Liu, Zhang, Liu, Zhang, Yu, Dang, Yang, Men, Huang, Ren, Ren, Zhou, and Lin}]{qwen2.5coder}
Binyuan Hui, Jian Yang, Zeyu Cui, Jiaxi Yang, Dayiheng Liu, Lei Zhang, Tianyu Liu, Jiajun Zhang, Bowen Yu, Kai Dang, An~Yang, Rui Men, Fei Huang, Xingzhang Ren, Xuancheng Ren, Jingren Zhou, and Junyang Lin. 2024.
\newblock Qwen2.5-coder technical report.
\newblock \emph{CoRR}, abs/2409.12186.

\bibitem[{Husain et~al.(2019)Husain, Wu, Gazit, Allamanis, and Brockschmidt}]{codesearchnet}
Hamel Husain, Ho{-}Hsiang Wu, Tiferet Gazit, Miltiadis Allamanis, and Marc Brockschmidt. 2019.
\newblock Codesearchnet challenge: Evaluating the state of semantic code search.
\newblock \emph{CoRR}, abs/1909.09436.

\bibitem[{Kim et~al.(2020)Kim, Kang, On, Heo, and Zhang}]{Kim2020HypergraphAN}
Eun-Sol Kim, Woo-Young Kang, Kyoung-Woon On, Yu-Jung Heo, and Byoung-Tak Zhang. 2020.
\newblock Hypergraph attention networks for multimodal learning.
\newblock In \emph{2020 {IEEE/CVF} Conference on Computer Vision and Pattern Recognition, {CVPR} 2020}, pages 14569--14578.

\bibitem[{Kingma and Ba(2015)}]{Kingma2015Adam}
Diederik~P. Kingma and Jimmy Ba. 2015.
\newblock Adam: {A} method for stochastic optimization.
\newblock In \emph{The 3rd International Conference on Learning Representations, {ICLR} 2015}.

\bibitem[{Li et~al.(2023)Li, Allal, Zi, Muennighoff, Kocetkov, Mou, Marone, Akiki, Li, Chim, Liu, Zheltonozhskii, Zhuo, Wang, Dehaene, Davaadorj, Lamy{-}Poirier, Monteiro, Shliazhko, Gontier, Meade, Zebaze, Yee, Umapathi, Zhu, Lipkin, Oblokulov, Wang, V, Stillerman, Patel, Abulkhanov, Zocca, Dey, Zhang, Fahmy, Bhattacharyya, Yu, Singh, Luccioni, Villegas, Kunakov, Zhdanov, Romero, Lee, Timor, Ding, Schlesinger, Schoelkopf, Ebert, Dao, Mishra, Gu, Robinson, Anderson, Dolan{-}Gavitt, Contractor, Reddy, Fried, Bahdanau, Jernite, Ferrandis, Hughes, Wolf, Guha, von Werra, and de~Vries}]{starcoder}
Raymond Li, Loubna~Ben Allal, Yangtian Zi, Niklas Muennighoff, Denis Kocetkov, Chenghao Mou, Marc Marone, Christopher Akiki, Jia Li, Jenny Chim, Qian Liu, Evgenii Zheltonozhskii, Terry~Yue Zhuo, Thomas Wang, Olivier Dehaene, Mishig Davaadorj, Joel Lamy{-}Poirier, Jo{\~{a}}o Monteiro, Oleh Shliazhko, and 48 others. 2023.
\newblock Starcoder: may the source be with you!
\newblock \emph{Trans. Mach. Learn. Res.}, 2023.

\bibitem[{Lin and Och(2004)}]{bleu}
Chin{-}Yew Lin and Franz~Josef Och. 2004.
\newblock {ORANGE:} a method for evaluating automatic evaluation metrics for machine translation.
\newblock In \emph{The 20th International Conference on Computational Linguistics, {COLING} 2004}.

\bibitem[{Liu et~al.(2019)Liu, Ott, Goyal, Du, Joshi, Chen, Levy, Lewis, Zettlemoyer, and Stoyanov}]{Liu2019RoBERTaAR}
Yinhan Liu, Myle Ott, Naman Goyal, Jingfei Du, Mandar Joshi, Danqi Chen, Omer Levy, Mike Lewis, Luke Zettlemoyer, and Veselin Stoyanov. 2019.
\newblock Roberta: A robustly optimized bert pretraining approach.
\newblock \emph{CoRR}, abs/1907.11692.

\bibitem[{Loshchilov and Hutter(2019)}]{LoshchilovHAdamW}
Ilya Loshchilov and Frank Hutter. 2019.
\newblock Decoupled weight decay regularization.
\newblock In \emph{The 7th International Conference on Learning Representations, {ICLR} 2019}.

\bibitem[{Lu et~al.(2021)Lu, Guo, Ren, Huang, Svyatkovskiy, Blanco, Clement, Drain, Jiang, Tang, Li, Zhou, Shou, Zhou, Tufano, Gong, Zhou, Duan, Sundaresan, Deng, Fu, and Liu}]{LuGRHSBCDJTLZSZ21}
Shuai Lu, Daya Guo, Shuo Ren, Junjie Huang, Alexey Svyatkovskiy, Ambrosio Blanco, Colin~B. Clement, Dawn Drain, Daxin Jiang, Duyu Tang, Ge~Li, Lidong Zhou, Linjun Shou, Long Zhou, Michele Tufano, Ming Gong, Ming Zhou, Nan Duan, Neel Sundaresan, and 3 others. 2021.
\newblock Codexglue: {A} machine learning benchmark dataset for code understanding and generation.
\newblock In \emph{Proceedings of the Neural Information Processing Systems Track on Datasets and Benchmarks 1, NeurIPS Datasets and Benchmarks 2021}.

\bibitem[{Montella et~al.(2023)Montella, Nasr, Heinecke, B{\'{e}}chet, and Rojas{-}Barahona}]{MontellaNHBR23}
S{\'{e}}bastien Montella, Alexis Nasr, Johannes Heinecke, Fr{\'{e}}d{\'{e}}ric B{\'{e}}chet, and Lina~Maria Rojas{-}Barahona. 2023.
\newblock Investigating the effect of relative positional embeddings on amr-to-text generation with structural adapters.
\newblock In \emph{Proceedings of the 17th Conference of the European Chapter of the Association for Computational Linguistics, {EACL} 2023}, pages 727--736.

\bibitem[{Niu et~al.(2022)Niu, Li, Luo, and Ng}]{NiuL0022}
Changan Niu, Chuanyi Li, Bin Luo, and Vincent Ng. 2022.
\newblock Deep learning meets software engineering: {A} survey on pre-trained models of source code.
\newblock In \emph{Proceedings of the 31st International Joint Conference on Artificial Intelligence, {IJCAI} 2022}, pages 5546--5555.

\bibitem[{Pfeiffer et~al.(2021)Pfeiffer, Kamath, R{\"u}ckl{\'e}, Cho, and Gurevych}]{Pfeiffer2020AdapterFusionNT}
Jonas Pfeiffer, Aishwarya Kamath, Andreas R{\"u}ckl{\'e}, Kyunghyun Cho, and Iryna Gurevych. 2021.
\newblock Adapterfusion: Non-destructive task composition for transfer learning.
\newblock In \emph{Proceedings of the 16th Conference of the European Chapter of the Association for Computational Linguistics: Main Volume, {EACL} 2021}, pages 487--503.

\bibitem[{Pfeiffer et~al.(2020)Pfeiffer, Vulic, Gurevych, and Ruder}]{Pfeiffer2020MADXAA}
Jonas Pfeiffer, Ivan Vulic, Iryna Gurevych, and Sebastian Ruder. 2020.
\newblock Mad-x: An adapter-based framework for multi-task cross-lingual transfer.
\newblock In \emph{Proceedings of the 2020 Conference on Empirical Methods in Natural Language Processing, {EMNLP} 2020}, pages 7654--7673.

\bibitem[{Radford and Narasimhan(2018)}]{Radford2018ImprovingLU}
Alec Radford and Karthik Narasimhan. 2018.
\newblock Improving language understanding by generative pre-training.

\bibitem[{Rebuffi et~al.(2017)Rebuffi, Bilen, and Vedaldi}]{Rebuffi2017LearningMV}
Sylvestre-Alvise Rebuffi, Hakan Bilen, and Andrea Vedaldi. 2017.
\newblock Learning multiple visual domains with residual adapters.
\newblock In \emph{Advances in Neural Information Processing Systems 30: Annual Conference on Neural Information Processing Systems 2017, {NeurIPS} 2017}, pages 506--516.

\bibitem[{Ribeiro et~al.(2021)Ribeiro, Zhang, and Gurevych}]{RibeiroZG21}
Leonardo F.~R. Ribeiro, Yue Zhang, and Iryna Gurevych. 2021.
\newblock Structural adapters in pretrained language models for amr-to-text generation.
\newblock In \emph{Proceedings of the 2021 Conference on Empirical Methods in Natural Language Processing, {EMNLP} 2021}, pages 4269--4282.

\bibitem[{Rozi{\`e}re et~al.(2023)Rozi{\`e}re, Gehring, Gloeckle, Sootla, Gat, Tan, Adi, Liu, Remez, Rapin et~al.}]{Rozire2023CodeLO}
Baptiste Rozi{\`e}re, Jonas Gehring, Fabian Gloeckle, Sten Sootla, Itai Gat, Xiaoqing Tan, Yossi Adi, Jingyu Liu, Tal Remez, J{\'e}r{\'e}my Rapin, and 1 others. 2023.
\newblock Code llama: Open foundation models for code.
\newblock \emph{CoRR}, abs/2308.12950.

\bibitem[{Svajlenko et~al.(2014)Svajlenko, Islam, Keivanloo, Roy, and Mia}]{svajlenko2014towards}
Jeffrey Svajlenko, Judith~F Islam, Iman Keivanloo, Chanchal~K Roy, and Mohammad~Mamun Mia. 2014.
\newblock Towards a big data curated benchmark of inter-project code clones.
\newblock In \emph{The 30th {IEEE} International Conference on Software Maintenance and Evolution, ICSME 2014}, pages 476--480.

\bibitem[{Vaswani et~al.(2017)Vaswani, Shazeer, Parmar, Uszkoreit, Jones, Gomez, Kaiser, and Polosukhin}]{Vaswani2017AAN}
Ashish Vaswani, Noam Shazeer, Niki Parmar, Jakob Uszkoreit, Llion Jones, Aidan~N. Gomez, Lukasz Kaiser, and Illia Polosukhin. 2017.
\newblock Attention is all you need.
\newblock In \emph{Advances in Neural Information Processing Systems 30: Annual Conference on Neural Information Processing Systems, {NeurIPS} 2017}, pages 5998--6008.

\bibitem[{Wang et~al.(2020)Wang, Li, Ma, Xia, and Jin}]{WangLM0J20}
Wenhan Wang, Ge~Li, Bo~Ma, Xin Xia, and Zhi Jin. 2020.
\newblock Detecting code clones with graph neural network and flow-augmented abstract syntax tree.
\newblock In \emph{The 27th {IEEE} International Conference on Software Analysis, Evolution and Reengineering, {SANER} 2020}, pages 261--271.

\bibitem[{Wang et~al.(2021)Wang, Wang, Joty, and Hoi}]{Wang2021CodeT5IU}
Yue Wang, Weishi Wang, Shafiq~R. Joty, and Steven C.~H. Hoi. 2021.
\newblock Codet5: Identifier-aware unified pre-trained encoder-decoder models for code understanding and generation.
\newblock In \emph{Proceedings of the 2021 Conference on Empirical Methods in Natural Language Processing, {EMNLP} 2021}, pages 8696--8708.

\bibitem[{Yadati et~al.(2019)Yadati, Nimishakavi, Yadav, Nitin, Louis, and Talukdar}]{Yadati2018HyperGCNAN}
Naganand Yadati, Madhav Nimishakavi, Prateek Yadav, Vikram Nitin, Anand Louis, and Partha~Pratim Talukdar. 2019.
\newblock Hypergcn: A new method for training graph convolutional networks on hypergraphs.
\newblock In \emph{Advances in Neural Information Processing Systems 32: Annual Conference on Neural Information Processing Systems 2019, {NeurIPS} 2019}, pages 1509--1520.

\bibitem[{Yang et~al.(2025)Yang, Li, Yang, Zhang, Hui, Zheng, Yu, Gao, Huang, Lv, Zheng, Liu, Zhou, Huang, Hu, Ge, Wei, Lin, Tang, Yang, Tu, Zhang, Yang, Yang, Zhou, Zhou, Lin, Dang, Bao, Yang, Yu, Deng, Li, Xue, Li, Zhang, Wang, Zhu, Men, Gao, Liu, Luo, Li, Tang, Yin, Ren, Wang, Zhang, Ren, Fan, Su, Zhang, Zhang, Wan, Liu, Wang, Cui, Zhang, Zhou, and Qiu}]{qwen3}
An~Yang, Anfeng Li, Baosong Yang, Beichen Zhang, Binyuan Hui, Bo~Zheng, Bowen Yu, Chang Gao, Chengen Huang, Chenxu Lv, Chujie Zheng, Dayiheng Liu, Fan Zhou, Fei Huang, Feng Hu, Hao Ge, Haoran Wei, Huan Lin, Jialong Tang, and 41 others. 2025.
\newblock Qwen3 technical report.
\newblock \emph{CoRR}, abs/2505.09388.

\bibitem[{Yang et~al.(2023)Yang, Jin, and Dou}]{YangJD23}
Guang Yang, Tiancheng Jin, and Liang Dou. 2023.
\newblock Heterogeneous directed hypergraph neural network over abstract syntax tree {(AST)} for code classification.
\newblock In \emph{The 35th International Conference on Software Engineering and Knowledge Engineering, {SEKE} 2023}, pages 274--279.

\bibitem[{Zhang et~al.(2024{\natexlab{a}})Zhang, Zeng, Wang, and Lu}]{tinyllama}
Peiyuan Zhang, Guangtao Zeng, Tianduo Wang, and Wei Lu. 2024{\natexlab{a}}.
\newblock Tinyllama: An open-source small language model.
\newblock \emph{CoRR}, abs/2401.02385.

\bibitem[{Zhang et~al.(2024{\natexlab{b}})Zhang, Chen, Liu, Liao, Gong, Yu, Li, and Wang}]{zhang2024unifying}
Ziyin Zhang, Chaoyu Chen, Bingchang Liu, Cong Liao, Zi~Gong, Hang Yu, Jianguo Li, and Rui Wang. 2024{\natexlab{b}}.
\newblock Unifying the perspectives of {NLP} and software engineering: A survey on language models for code.
\newblock \emph{Trans. Mach. Learn. Res.}, 2024.

\end{thebibliography}

\end{document}